\documentclass[twoside,11pt]{article}
\usepackage{pgm}

\usepackage[utf8]{inputenc}
\inputencoding{utf8}

\usepackage{hyperref,color,soul,booktabs}
\usepackage[american]{babel}
\usepackage{multirow}
\usepackage{hyperref}
\usepackage{tabularx}
\usepackage{booktabs}
\usepackage{graphicx}
\usepackage[dvipsnames]{xcolor}
\usepackage{amsmath,amsfonts}
\usepackage{wrapfig}
\usepackage{algorithm}
\usepackage[format=plain]{caption}
\usepackage[algo2e,ruled,linesnumbered]{algorithm2e} 
\usepackage{algorithmic}
\usepackage{xcolor}
\usepackage{enumitem}
\usepackage{mathtools}
\usepackage{tikz}
\usetikzlibrary{arrows}
\tikzset{
  treenode/.style = {align=center, inner sep=0pt, text centered, font=\sffamily},
  product_node/.style = {treenode, circle, black, draw=none,
    fill=blue!20, text width=1.5em},  
  sum_node/.style = {treenode, circle, black,  draw=none, 
    fill=red!20, text width=1.5em},
  leaf_node/.style = {treenode, circle, black, draw=none,
    fill=green!20, text width=1.5em}
}

\newcommand{\X}{\mathbf{X}}

\newcommand{\w}{\mathbf{w}}
\newcommand{\N}{\mathbf{N}}
\newcommand{\E}{\mathbf{E}}
\newcommand{\y}{\mathbf{y}}
\newcommand{\da}{\ensuremath{\mathcal{D}}}
\newcommand{\ind}{\perp\!\!\!\!\perp}
\newcommand{\tbu}[1]{\textbf{\underline{#1}}}
\newcommand{\ex}{\ensuremath{\mathcal{EXSPN}}}

\setulcolor{blue}
\newcommand{\BibTeX}{\textsc{B\kern-0.1emi\kern-0.017emb}\kern-0.15em\TeX}

\newlength{\bibitemsep}
\newlength{\bibparskip}
\let\oldthebibliography\thebibliography
\renewcommand\thebibliography[1]{%
  \oldthebibliography{#1}%
  \setlength{\parskip}{\bibitemsep}%
  \setlength{\itemsep}{\bibparskip}%
}

\ShortHeadings{\ex:Explaining Sum-Product Networks}{Karanam et al.}

\begin{document}

\title{Explaining Deep Tractable Probabilistic Models: \\ The sum-product network case}

\author{\Name{Athresh Karanam\footnotemark[1]} \Email{athresh.karanam@utdallas.edu} \and
    \Name{Saurabh Mathur\footnotemark[1]} \Email{saurabhsanjay.mathur@utdallas.edu}\and
    \Name{Predrag Radivojac} \Email{predrag@northeastern.edu}\and
    \Name{David M. Haas} \Email{dahaas@iu.edu}\and
    \Name{Kristian Kersting} \Email{kersting@cs.tu-darmstadt.de}\and
    \Name{Sriraam Natarajan} \Email{sriraam.natarajan@utdallas.edu}
}
\maketitle

\footnotetext[1]{*Equal contributors}

\begin{abstract}
We consider the problem of explaining a class of tractable deep probabilistic models, the Sum-Product Networks (SPNs) and present an algorithm \ex to generate explanations. To this effect, we define the notion of a context-specific independence tree(CSI-tree) and present an iterative algorithm that converts an SPN to a CSI-tree. The resulting CSI-tree is both interpretable and explainable to the domain expert. We achieve this by extracting the conditional independencies encoded by the SPN and approximating the local context specified by the structure of the SPN. Our extensive empirical evaluations on synthetic, standard, and real-world clinical data sets demonstrate that the CSI-tree exhibits superior explainability.
\end{abstract}

\section{Introduction}\label{sec:intro}

{\bf Tractable Deep Probabilistic Models} (TDPMs) exploit the efficiency of deep learning (\cite{DeepLearningBook2016}) while abstracting the representation of the underlying model. TDPMs implement {\bf compositions of functions}, which increases their representation power considerably compared to deep learning. Specifically, they abstract the underlying representation by implementing a composition of probability distributions over domain features, which can be discrete, continuous,  graphical, or even unstructured. Unsurprisingly, these benefits have led to considerable interest in TDPMs~(\cite{choi2017relaxing, poon2011sum, CutajarEtAl17}). Some TDPMs such as Arithmetic Circuits and Sum-product Networks (SPNs) explicitly model the joint distribution using a {\bf network polynomial} over evidence indicators and network parameters. This makes them closely related to polynomial neural networks (PNNs, \cite{NikolaevHitoshi06}), which are a class of power-series function models with multiplicative activation functions and parsimonious structure.

We consider the specific formulation of SPNs and pose the following question -- {\em can SPNs with their multiple layers be explained using existing tools inside probabilistic modeling?} To achieve this, we move beyond the traditional notions of conditional independencies that can be read off an SPN and instead focus on context-specific independencies (CSI, \cite{boutilier}). CSIs provide a more in-depth look at the relationships between two variables when affecting the third variable. For instance, in a gestational diabetes prediction task (\cite{karanamEtAl2021}), one could state that gestational diabetes and education are conditionally independent given the age. This allows the care provider/physician to develop a good interventional treatment plan. Recent work on developing interventions given a learned SPN (\cite{iSPN2021}) demonstrates the potential of such TDPMs and our work goes in the same direction by identifying CSIs that could potentially aid the expert in identifying appropriate interventions. 

Specifically, we define the notion of a CSI-tree that is used as a {\em visual tool to explain SPNs}. We present an algorithm (\ex) that grows a CSI-tree iteratively. We show clearly that the  constructed CSI-tree can recover the full SPN structure. Once a tree is constructed, we then approximate the CSIs by learning a supervised model to fit the CSIs. The resulting feature importances can then be used to further approximate the tree. Our evaluations against association rule mining clearly demonstrate that the recovered CSI-trees indeed are shorter and more interpretable.
\begin{wrapfigure}{R}{0.3\textwidth}
    \centering
    \scalebox{1.5}{
        \begin{tikzpicture}[->,>=stealth',level/.style={level distance = 1cm}, level 1/.style={sibling distance=1.5cm}, level 2/.style={sibling distance=0.8cm}, level 4/.style={sibling distance=0.8cm}] 
        \tikzstyle{edge from parent}=[draw,black]
        \node [sum_node] {\tiny $+$} 
            child{ node [product_node] {\tiny $\times$}  
                    child{ node [leaf_node] {\tiny $X_1$}}
                    child{ node [leaf_node] {\tiny $X_2$}}
                    edge from parent node[sloped, auto=left] {\tiny \sffamily 0.3}
            } 
           child{ node [product_node] {\tiny $\times$}  
                    child{ node [leaf_node] {\tiny $X_1$} } 
                    child{ node [leaf_node] {\tiny $X_2$} }
                    edge from parent node[sloped, auto=left] {\tiny \sffamily 0.7}
            } 
        
        ; 
        \end{tikzpicture}
    }
    \caption{\small SPN that represents the joint distribution $P(X_1, X_2)$. Sum nodes are inscribed with ''+'', product nodes with ''$\times$'' and leaf nodes with corresponding variable names(\em best viewed in color).}\label{fig:simple_spn}
    \vspace{-1.5em}
\end{wrapfigure}
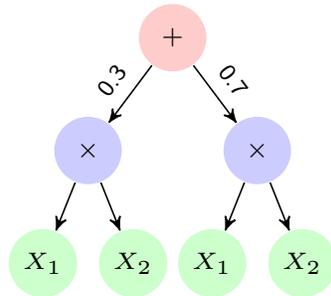

We make the following key contributions: (1) We develop CSI-trees that focus on the explanation. These trees are built on the earlier successes inside graphical models and we use them in the context of  TDPMs.  (2) We develop an iterative procedure that constructs a CSI-tree given an SPN. The resulting tree is a complete representation of the original SPN. We present an approximation heuristic that compresses these trees further to enhance the explainability of the model. One key aspect of \ex is that it is {\bf independent} of the underlying SPN learning algorithm. Any SPN that is complete and consistent (as defined in the next section) can be used as input for \ex.~\footnote {Code and appendix is available at \texttt{\small{anonymous.4open.science/r/ExSPN-16FD}}} (3) We perform extensive experiments on synthetic, multiple standard data sets and a real clinical data set. Our evaluation demonstrates that the final model induces smaller rules/models compared to the original SPN. 


\vspace{-0.5em}
\section{Background and Preliminaries}
\textbf{Sum-product networks:} SPNs(\cite{poon2011sum}) are weighted directed acyclic graphs(DAGs) with sum and product nodes as the internal nodes of the graph and probability distributions as the leaf nodes. They represent joint distributions over a set of variables $\X$. 
Let $\X = \{X_1, X_2,..., X_m\}$ be the set of $m$ random variables for which $|\da|$ data samples are available. We use the following definitions from ~\cite{zhaoEtAl2015}. 
A {\bf computational graph} $G$ is a DAG with three node types: {\bf sums}, {\bf products} and {\bf leaves}. $N$ denotes a generic node and $\N$ denotes a set of nodes in $G$. Similarly, $e$ denotes an edge and $\E$ denotes a set of edges in $G$. A {\bf scope function} $\psi$: $\N \rightarrow 2^{\X}$ is a mapping between a node $N$ and a subset of $\X$. It confines the set of RVs that a node is defined over. 
An {\bf SPN} $S$ is a 4-tuple $(G, \psi, \w, \theta)$, where $G$ is a computational graph, $\psi$ is a $scope$ $function$, $\w$ is a set of sum-weights and $\theta$ is a set of parameters of the leaf distributions.  $ch(N)$ denote the children of a node $N$ and $pa(N)$, its parents.

\textbf{SPN properties}: An SPN is \emph{complete} iff each sum node has children with the same scope. An SPN is \emph{consistent} iff no variable appears negated in one child of a product node and non-negated in another. An SPN is \emph{decomposable} iff for every product node $N$, the scope of its children are disjoint. An SPN is said to be {\bf normal} if (1) It is complete and decomposable, (2) For each sum node, the weights of the edges emanating from the sum node are nonnegative and sum to 1, and (3) Every terminal node in the SPN is a univariate distribution over a Boolean variable and the size of the scope of a sum node is at least 2. An \textbf{instance function} $\phi_{\da}: N \rightarrow 2^{|\da|}$ is a mapping between a node $N$ and a subset of $\da$. For notational simplicity, we use $\phi$ instead of $\phi_{\da}$ when $\da$ is implied.

\noindent \textbf{Example:} Figure \ref{fig:simple_spn} shows an SPN defined over two random variables $X_1, X_2$. The nodes of the graph with \colorbox{red!20}{``$+$''} within them are sum nodes, those with \colorbox{blue!20}{``$\times$''} within them are product nodes and the rest of the nodes are \colorbox{green!20}{leaf nodes}. The variable name within each leaf node indicates that a univariate distribution over that variable is learnt at that node. The set of sum, product and leaf nodes, and the edges connecting them constitute the SPN's computational graph. The scope of the sum node at the root and the two product nodes is $X_1,X_2$. The scope of the leaf nodes from left to right is $X_1$, $X_2$, $X_1$ and $X_2$, respectively. The labels on the edges from sum node to product nodes are the sum-weights. 

\noindent\textbf{A note on interpretability}: There is no unique definition of interpretability (\cite{Lipton2018MythosOfModelInterpretability},\cite{DoshiVelez2017TowardsAR}). In addition, it is often used interchangeably with explanability, with some work making a distinction (\cite{MontavonSM17}, \cite{Rudin19}). In this work, by interpretable models, we mean representations whose random variables, dependencies(structure) and parameters are interpretable by humans (\cite{TowellShavlik1991KBANNtoRules}, \cite{MontavonSM17}). CSI-trees do not introduce any latent variables, unlike SPNs(\cite{PeharzGPD16}), and their parameters are logical statements comprising observed variables, thus satisfying the criterion for interpretability.

\textbf{Learning SPNs:} Several learning algorithms(\cite{DBLP:conf/icml/GensD13}, \cite{DBLP:conf/aaai/0001VMNEK18}) have been proposed for SPNs. For brevity, we will limit our discussion to simultaneous parameter and structure learning for tree-structured SPNs using the popular learnSPN (\cite{DBLP:conf/icml/GensD13}) framework. 
Each recursive step of their algorithm either learns parameters of a leaf distribution, a sum node by splitting the data instances into subsets, or a product node by decomposition of the variables into subsets of mutually independent variables. In the base case, when conditions for learning the leaf distributions are satisfied, a univariate leaf distribution is learnt and the recursion ends. If the variables can be partitioned into mutually independent subsets $\X_i \subseteq \X$, the algorithm learns a product node and recurses over each subset. Otherwise, the data is partitioned into subsets $\da_j \subset \da$, the algorithm learns a sum node and recurses over each subset. 
The sum and product nodes of SPNs learnt using learnSPN are latent variables whereas the leaf nodes learn distributions over observed variables $\X$. 
\vspace{-1em}


\section{\ex - Explaning SPNs}

\begin{figure*}
    \centering
    \includegraphics[width=\textwidth, height = 0.3\textwidth]{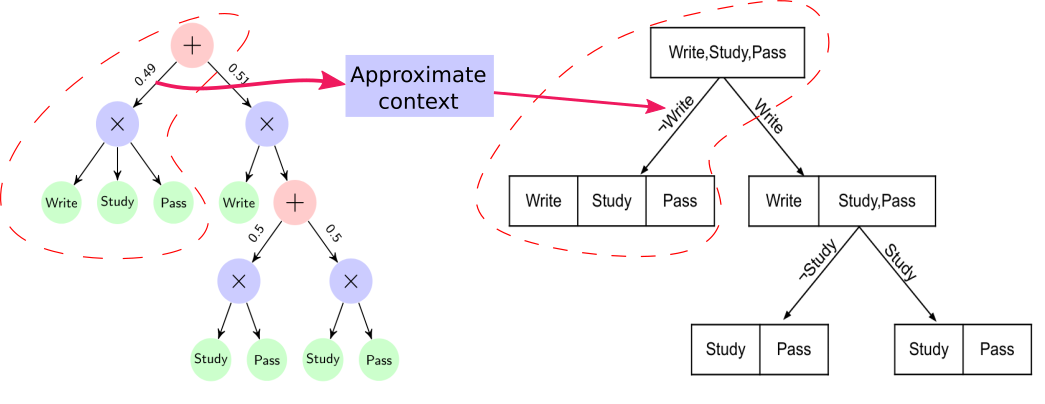}
    \caption{\small A normal SPN (left) and its corresponding CSI-Tree (right). The independencies induced by the product node are represented in the partitioned leaf node of the CSI-tree. The context in which these independencies hold is approximated to $\neg Write$. }
    \label{fig:example_spn}
    \vspace{-1em}
\end{figure*}
Before we outline our procedure for explaining SPNs, we briefly explain the notion of \noindent Context-specific independence(CSI) (\cite{boutilier}). CSI is a generalization of the concept of statistical independence of random variables. CSI-relations have been studied extensively over the last three decades (\cite{boutilier,nyman2014stratified,DBLP:conf/nips/TikkaHK19,nyman2016context}). They can be used to speed-up probabilistic inference, improve structure learning and explain graphical models that are learnt from data. We propose {\bf a novel algorithm to extract these CSI-relations from SPNs} and empirically demonstrate that the extracted CSI-relations are interpretable. CSIs can be directly extracted from the data by computing conditional probabilities (\cite{shen2020new}) with context using parameterized functions such as neural networks(\cite{kingma2013auto}). While the literature on rule learning is vast (\cite{FurnkranzEtAl2015RuleLearning}), we note that our objective is \underline{\emph{explaining SPNs} and not \emph{rule learning}}. Additionally, we note that the relationship between SPNs and BNs (\cite{zhaoEtAl2015}), and that between SPNs and multi-layer perceptrons (\cite{VergariEtAl2019}) is well established. However, unlike our work, these methods do not explicitly attempt to explain SPNs through a compact representation of CSIs. The algorithm to convert SPNs to BNs presented by Zhao et. al.(\cite{zhaoEtAl2015}) introduces unobserved hidden variables into the BN. It generates a BN with a directed bipartite structure with a layer of hidden variables pointing to a layer of observable variables. In this case, the BN associates each sum node in the SPN to a hidden variable in BN. We argue that the introduction of these hidden variables renders the resulting BN uninterpretable.

To construct meaningful explanations, we define a compact and interpretable representation of CSIs called {\bf CSI-tree}. 
\vspace{-0.75em}
\begin{Definition}
    (CSI-tree): A CSI-tree $\tau$ is a 4-tuple $(G, \psi_{csi}, \chi, \zeta)$ where $G$ is a tree with a set of variables $V \subseteq \X$, \emph{scope function} $\psi_{csi}$, \emph{partition function} $\chi$ and edge labels $\zeta$.
\end{Definition}
\vspace{-1em}
\begin{Definition}
    (Partition function): A partition function $\chi_{\psi}: N \rightarrow 2^{|\psi(N)|}$ is a mapping from a node to a set $C$ of disjoint sets $P_i \subseteq \psi(N)$ under a given scope function $\psi$ such that $\cup_{i \in |C|} P_i = \psi(N)$.
\end{Definition}
\vspace{-0.75em}

Each node, $N$, of a CSI-tree has a scope $\psi_{csi}(N) \subseteq \X$. The partition function divides the variables within the scope of each node into disjoint subsets such that the union of these subsets is $\psi_{csi}(N)$. The edges are labeled with a conjunction $\zeta$ over a subset of $\X$. An example of an edge label is ($X_i \geq 0.5 \wedge X_j \leq 1$). Essentially, the edge label narrows the scope of the context from parent to child node. 

\noindent \textbf{Example CSI-tree:} The right side of the Figure \ref{fig:example_spn} shows a CSI-tree defined over the binary variables $\langle Write,Study,Pass \rangle$. The set of variables inscribed within each node represent the scope of that node. For example, the scope of the root node is $\langle Write,Study,Pass \rangle$. The edge label \emph{confines the context} by conditioning on a set of variables (on a singleton set in this example). The variables in the scope of a node that are conditionally independent when conditioned on the context are separated into subsets (denoted by a vertical bar). It is the output of the partition function for that node. For example, the right child of the root node specifies a partition $\langle \langle Write \rangle, \langle Study, Pass \rangle \rangle$ of the scope of that node. 

The left child of the root node induces a CSI-relation $Write \ind Study \ind Pass \vert \neg Write$. Similarly, the right child of the root node induces two CSI-relations $Write \ind Study \vert Write$ and $Write \ind Pass \vert Write$. To summarize, the subsets of variables within the scope of a node, as defined by the partition function, are independent of each other when conditioned on the proposition specified in the label of the edge connecting that node to its parent. Note that {\em reading this CSI-tree is significantly easier than a SPN} on the left, due to the internal nodes entirely comprising of observed variables, and thus allows for a more explainable and interpretable representation. Now, we formally define our goal:


\fbox{\centering\begin{minipage}{20em}
    
    \textbf{Given:} SPN $S=(G, \psi, w, \theta)$, data $\da$\\
    \textbf{To Do:} Extract $\text{CSI-tree}$ $\tau=(G_{csi}, \psi_{csi}, \chi, \zeta)$
    \end{minipage}
}

\noindent The left side of Figure \ref{fig:example_spn} shows an SPN learnt over the variables $Write, Study, Pass$. The root node $N_0$ is a sum node, implying that $Write, Study, Pass$ are not independent of each other in the context of the entire dataset. Now, consider the left child of the root node $N_1$, which is a product node with three leaf nodes as its children. It implies that $Write \ind Study \ind Pass \vert \phi(N_1)$. In other words, the $Write, Study, Pass$ are {\bf independent of each other in the context of $\phi(N_1)$}. While the context of $\phi(N_1)$ accurately explains the conditional independence induced by the product node, it requires $2^{|\psi(N_1)|}$ parameters to fully parameterize the context which renders the CSI specified under $\phi(N_1)$ uninterpretable. Therefore, we need to approximate this context in order to ensure interpretability of these CSI-relations.
Additionally, while the instantiation of a subset of $\psi(N_1)$ that is consistent with $\phi(N_1)$ can be used to define the context for a particular CSI, this method would be highly sensitive to noise. 

To address these issues, we propose to {\bf first learn a discriminative model $f$} in the supervised learning paradigm to predict if a data point belongs to $\phi(N_1)$ and use the notion of {\bf feature importance to approximate the context}. We identify a set $I \subseteq \psi(N_1)$ of the most important features as determined by the feature importance scores for $\psi(N_1)$ w.r.t $f$ and approximate the context to a proposition of the form $\wedge_{i \in I}$. For this example, we choose $f$ to be a \emph{decision tree} and the \emph{mean decrease in impurity} as the measure of feature importance and obtain $(\neg Write)$ as the approximated context. So, the original CSI $Write \ind Study \ind Pass \vert \phi(N_1)$ is approximated to $Write \ind Study \ind Pass \vert \neg Write$.
\vspace{-1em}
\subsection{\ex Algorithm}
\begin{wrapfigure}{L}{0.5\textwidth}
    \begin{minipage}{0.5\textwidth}
    \vspace{-1em}
    \begin{algorithm2e}[H]
    \small
      \SetAlgoLined
      \caption{\ex}
      \label{alg:ExSPN}
      \SetKwInOut{Input}{input}\SetKwInOut{Output}{output}
      \Input{$\da$, $\X$, $S=(G=(\N, \E), \psi, w, \theta)$, $\lambda$}
      \Output{CSI-tree $\tau=(G_{csi}, \psi_{csi}, \chi, \zeta)$}
      Convert $S$ to a normal-spn $S_{normal}$\\
      Infer $\phi$ using alg. \ref{alg:InferInstanceFunction}\\
      Initialize $G_{csi}=G_{normal}, \psi_{csi}=\psi_{normal}$, \\ $\tau=(G_{csi}, \psi_{csi}, \chi, \zeta)$ \\ 
      st = [$G_{csi}$.root] \\
      Add $\psi(G_{csi}.root)$ to $\chi(G_{csi}.root)$ \\
      \While{$st$ is not empty}{
      $N_{current}=st.pop()$ \\
      \If{$N_{current}$ is not the root node}{
      \If{$N_{current}$ is a leaf node}{
      Add $\psi(N_{current})$ to $\chi(pa(N_{current}))$
      }
      \If{$N_{current}$ is a sum node}{
      Add $\psi(N_{current})$ to $\chi(pa(N_{current}))$ \\
      Connect $pa(N_{current})$ and each $ch(N_{current})$\\
      Replicate sub-SPN rooted at $ch(N_{current})$ 
      for each $|pa(ch(N_{current}))|$ to ensure  $G_{csi}$ has a tree structure \\
      Delete $N_{current}$ from $G_{csi}$
      }
      
      }
      Add $ch(N_{current})$ to $st$
      }
      $\tau=ComputeLabels(\tau, \da, \phi, \lambda)$ \\
      \textbf{return} $\tau$
    \end{algorithm2e}
    \vspace{-0.5em}
    \end{minipage}
\end{wrapfigure}
For simplicity, we present our approach with binary variables. However, in our experiments, we demonstrate that our method {\bf can be applied to continuous and multi-class discrete variables} as well. We now outline our approach to infer CSI-trees from SPNs. The steps involved in converting an SPN into a CSI-tree are:
(1) Convert $S$ into a normal-SPN $S_{normal}$. (2) Infer the instance function $\phi$. (3) Create a CSI-tree $\tau_{unlabeled}$ with no edge labels using $\phi$. (4) Compute edge labels for $\tau_{unlabeled}$, and create CSI-tree $\tau$. (5) Optionally, compress $\tau$ into $\tau_{compressed}$ by pruning $\tau$.


Algorithm \ref{alg:ExSPN} presents \ex: \tbu{Ex}plaining \tbu{S}um-\tbu{P}roduct \tbu{N}etworks. It infers a CSI-tree $\tau$, given an SPN $S$, data $\da$, set of random variables $\X$ and feature importance score threshold $\lambda$.

\noindent(\textbf{Step 1:}) Convert $S$ to a normal-spn $S_{normal}=(G_{normal}, \psi_{normal}, w_{normal}, \theta_{normal})$ \\ \textbf{[line 1]} using the conversion scheme proposed by Zhao et. al. (\cite{zhaoEtAl2015}). Any arbitrary SPN $S$ can be converted into a normal SPN $S_{normal}$ that represents the same joint probability over variables and $|S_{normal}|=\mathcal{O}(|S|^2)$.

\noindent(\textbf{Step 2:}) It then infers the instance function for $S_{normal}$\textbf{[line 2]} using algorithm \ref{alg:InferInstanceFunction}. This algorithm is similar to the algorithm proposed in \cite{poon2011sum} for approximating most probable explanation(MPE) inference in arbitrary SPNs. For each instance $\da_j \in \da$ it first performs an upward pass from leaf nodes to the root node and computes $S_i^{max}(\da_j)$ for each node $N_i$ as follows:
\begin{itemize}
    \item if $N_i$ is a sum node, then, $S_i^{max}(\da_j) = \max_{k\in ch(N_i)}w_{ik}.S_j^{max}(\da_j)$
    \item otherwise $S_i^{max}(\da_j)=S_i(\da_j)$ 
\end{itemize}
Then the algorithm backtracks from the root to the leaves, appending $\da_j$ to the instance function associated with each child of a sum node that led to $S_i^{max}(\da_j)$ and to the instance function associated with all product nodes along the path from the root to a leaf node.

\noindent(\textbf{Step 3:}) Next, it performs a depth-first search(DFS) on the computational graph $G_{normal}$ associated with $S_{normal}$ and constructs the tree-structured graph $G_{csi}$ associated with $\tau$ \textbf{[lines 4-19]}. $G_{csi}$ is initialized with $G$ and the scope of root of $G_{csi}$, $\psi(G_{csi}.root)$ is added to the value of the partition function associated with $\tau$\textbf{[line 5]}. \ex maintains a stack $st$ that is initialized with the root of $G_{csi}$\textbf{[line 4]}. The current node selected in DFS, $N_{current}$, is popped from the stack $st$\textbf{[line 7]}. It then considers three cases: 1. $N_{current}$ is a leaf node 2. $N_{current}$ is a sum node 3. $N_{current}$ is a product node\textbf{[lines 9-16]}. If $N_{current}$ is a leaf node, its scope $\psi(N_{current})$ is added to the partition function of its parent node $\chi(pa(N_{current}))$\textbf{[lines 9-10]}. If $N_{current}$ is a sum node, its scope $\psi(N_{current})$ is added to the partition function of its parent node $\chi(pa(N_{current}))$, edges are added to $G_{csi}$ to connect the parent of $N_{current}$, $pa(N_{current})$, to each child of $N_{current}$, and deleting $N_{current}$ and all edges $e$ in $G_{csi}$ of the form $e(\alpha, N_{current})$ or $e(N_{current}, \alpha)$\textbf{[line 12-16]}. If $N_{current}$ is a product node, it is ignored. It then continues DFS over $G_{csi}$ by adding all the children of $N_{current}$ to $st$\textbf{[line 19]}. Note that the CSI-tree $\tau$ is unlabeled.

\begin{wrapfigure}{R}{0.5\textwidth}
    \begin{minipage}{0.5\textwidth}
        \vspace{-2em}
        \begin{algorithm2e}[H]
        \small
            \SetAlgoLined
            \caption{\small Infer Instance Function}
            \label{alg:InferInstanceFunction}
            \SetKwInOut{Input}{input}\SetKwInOut{Output}{output}
            \Input{$\da$, $\X$, normal SPN $S$}
            \Output{Instance function $\phi$}
            \For{Each instance $\da_j \in \da$}{
            Perform upward pass for $\mathcal{D}$  \\
            Compute $S_N^{max}(\da_j)$ for each node $N$ \\
            Perform a downward pass \\
            Append $\da_j$ to the $\phi(N_{ch})$ of a child $N_{ch}$ of a sum node that led to $S^{max}(\da_j)$ \\
            Append $\da_j$ to $\phi(N_{product})$ for all product nodes $N_{product}$ along the path from the root to a leaf node \\
            }
            \textbf{return} $\phi$
        \end{algorithm2e}
        \vspace{-1.5em}
    \end{minipage}
\end{wrapfigure}
\noindent(\textbf{Step 4:}) Algorithm \ref{alg:ComputeLabels} in the Appendix presents the procedure to train a model $f$, compute a set of important features $I$, and finally compute the edge labels $\zeta$. For each edge $e(N_{from}, N_{to})$, it first computes class labels $y$ to indicate if an instance $\da_j\in \phi(N_{from})$ belongs to $\phi(N_{to})$. Then it computes a set of important features $I$ for $f$ using a suitable feature importance computation technique based on the choice of $f$ and a threshold for feature importance score $\lambda$. The edge label for $e$ is the conjunction of the features in $I$.

\noindent(\textbf{Step 5:}) The labeled CSI-tree $\tau$ created in the previous step might be too large in some cases and the CSIs induced by $\tau$ at a product node $N$ may be supported by a small number of examples given by $|\phi(N)|$. To avoid these two issues, we propose deleting the sub-tree induced by a product node $N$ for which $|\phi(N)|< min_{instances}$. This heuristic significantly reduces the size of the CSI-tree while retaining CSIs induced by product nodes closer to the root node of the SPN, as demonstrated in our experiments.

    


\noindent \textbf{Computational Complexity}: The computational complexity of \ex algorithm depends on the complexity of each of its constituent 5 steps. The size ($\#nodes+\#edges$) of the normal-SPN $S'$ obtained from the original SPN $S$ has a space-complexity of $O(|S|^2)$. This conversion can be done is time linear in the size of the normal-SPN (\cite{zhaoEtAl2015}). The generation of instance function for the CSI-tree involves MAX inference for each of $\da$ data points which can be performed in $O(|\da||S'|)$ (\cite{poon2011sum}). The generation of unlabeled CSI-tree has a time-complexity of $O(|S'|)$. Generating edge labels involves training a discriminative model which can be performed in $O(|\da||\theta_{D}|)$ for a universal approximator such as neural network with parameters $\theta_D$. Here $|N_{sum}|$ is the number of sum nodes in the network and $|\mathbf{X}|$ is the number of variables. Finally the compression can be performed in time linear in the size of the network. This gives us an \emph{overall time complexity of $O(max(|\theta|, |X|)|S|^2|\da|)$}.

\noindent \textbf{Properties of CSI-tree:} $\tau=(G_{csi}, \psi_{csi}, \chi, \zeta)$, corresponding to an SPN $S, \psi, w, \theta)$ obtained through \ex has the following properties: (1) Num nodes in $G_{csi}$ = num product nodes in $G_{normal}$ + 1. (2) The context induced by a product node $N_p$ in $S$ requires $2^{|\psi(N_p)|}$ parameters to be sufficiently expressed, while the approximate context from \ex has $<|\psi(N_p)|$ parameters. Additionally, the structure of a tree-structured normal-SPN can be retrieved from its corresponding CSI-tree in time linear in the size of the SPN. \vspace{-0.5em}
\begin{theorem}
 The CSI-tree $\tau$, inferred from an SPN, $S_{normal}$, using \ex can infer $G'_{normal}$, and $\psi'_{normal}$ which encodes the same CSIs as $S_{normal}$.
\end{theorem}
We present the proof in Appendix \ref{app:proofs}. 

\section{Experimental Evaluation}
We explicitly answer the following questions: ({\bf Q1: Correctness}) Does \ex recover all the CSIs encoded in an SPN? ({\bf Q2: Compression}) Can the CSIs be compressed further? ({\bf Q3. Baseline}) How do the CSIs extracted using \ex compare with a strong rule learner?  ({\bf Q4. Real data}) Does \ex extract reasonable CSIs in a real clinical study?
\textbf{System:} We implemented \ex using SPFlow library (\cite{Molina2019SPFlow}). We assume that the instance function is computed during the training process. For experiments where the instance function is computed separately after training, see Table \ref{tab:csi_i}. 
Since Decision Trees can be represented as a set of decision rules, we used the Classification and Regression Trees (CART, \cite{cart84}) algorithm as the explainable function approximator. We used scikit-learn's DecisionTreeClassifier (\cite{scikit-learn}) to implement CART. The hyperparameter configuration of these algorithms is shown in Table \ref{tab:hyperparams} in the appendix.


\textbf{Baseline:} To evaluate the CSIs extracted by \ex, we compared them with the association rules mined using the Apriori algorithm (\cite{apriori}). We used the Mlxtend library (\cite{raschkas_2018_mlxtend}) to implement this baseline. Since the Apriori algorithm requires binary features, we discretized the continuous variables in the datasets into 5 categories and one-hot encoded the categorical variables.

\begin{figure*}[!ht]
    \scalebox{1.2}{
        \begin{tikzpicture}[->,>=stealth',level/.style={level distance = 0.75cm, sibling distance = 1cm}, level 1/.style={sibling distance = 4cm}]
        \tikzstyle{edge from parent}=[draw,black]
        \node [sum_node] {\tiny $+$} 
            child{ node [product_node] {\tiny $\times$}  
                    child{ node [leaf_node] {\tiny V0}}
                    child[sibling distance = 2cm]{ node [sum_node]{\tiny $+$} 
                        child[sibling distance = 3cm]{ node [sum_node]{\tiny $+$}
                            child[sibling distance = 2.5cm]{ node [product_node] {\tiny $\times$}  
                                child[sibling distance = .75cm]{ node [leaf_node] {\tiny V1}}
                                child[sibling distance = .75cm]{ node [leaf_node] {\tiny V2}}
                                child[sibling distance = .75cm]{ node [leaf_node] {\tiny V3}}
                                edge from parent node[sloped,auto=left] {\tiny \sffamily 0.49}
                            }
                            child[sibling distance = 2.5cm]{ node [product_node] {\tiny $\times$}  
                                child[sibling distance = .75cm]{ node [leaf_node] {\tiny V1}}
                                child[sibling distance = .75cm]{ node [leaf_node] {\tiny V2}}
                                child[sibling distance = .75cm]{ node [leaf_node] {\tiny V3}}
                                edge from parent node[sloped,auto=left] {\tiny \sffamily 0.51}
                            }
                            edge from parent node[sloped,auto=left] {\tiny \sffamily 0.5}
                        }
                        child[sibling distance = 3cm]{ node [product_node] {\tiny $\times$}
                            child[sibling distance = 1cm]{ node [leaf_node] {\tiny V1}}
                            child{ node [sum_node]{\tiny $+$}
                                child[sibling distance = 1.5cm]{ node [product_node] {\tiny $\times$}  
                                    child[sibling distance = .75cm]{ node [leaf_node] {\tiny V2}}
                                    child[sibling distance = .75cm]{ node [leaf_node] {\tiny V3}}
                                    edge from parent node[sloped,auto=left] {\tiny \sffamily 0.52}
                                }
                                child[sibling distance = 1.5cm]{ node [product_node] {\tiny $\times$}  
                                    child[sibling distance = .75cm]{ node [leaf_node] {\tiny V2}}
                                    child[sibling distance = .75cm]{ node [leaf_node] {\tiny V3}}
                                    edge from parent node[sloped,auto=left] {\tiny \sffamily 0.48}
                                }
                            }
                            edge from parent node[sloped,auto=left] {\tiny \sffamily 0.5}
                        }
                    }
                    edge from parent node[sloped,auto=left] {\tiny \sffamily 0.67}
            } 
           child[sibling distance = 3cm]{ node [product_node] {\tiny $\times$}  
                    child[sibling distance = .75cm]{ node [leaf_node] {\tiny V0} } 
                    child[sibling distance = .75cm]{ node [leaf_node] {\tiny V1} }
                    child[sibling distance = .75cm]{ node [leaf_node] {\tiny V2} } 
                    child[sibling distance = .75cm]{ node [leaf_node] {\tiny V3} }
                    edge from parent node[sloped,auto=left] {\tiny \sffamily 0.33}
            } 
        
        ; 
        \end{tikzpicture}
    
    }\hspace{-2.5em}
    \includegraphics[width=0.5\textwidth]{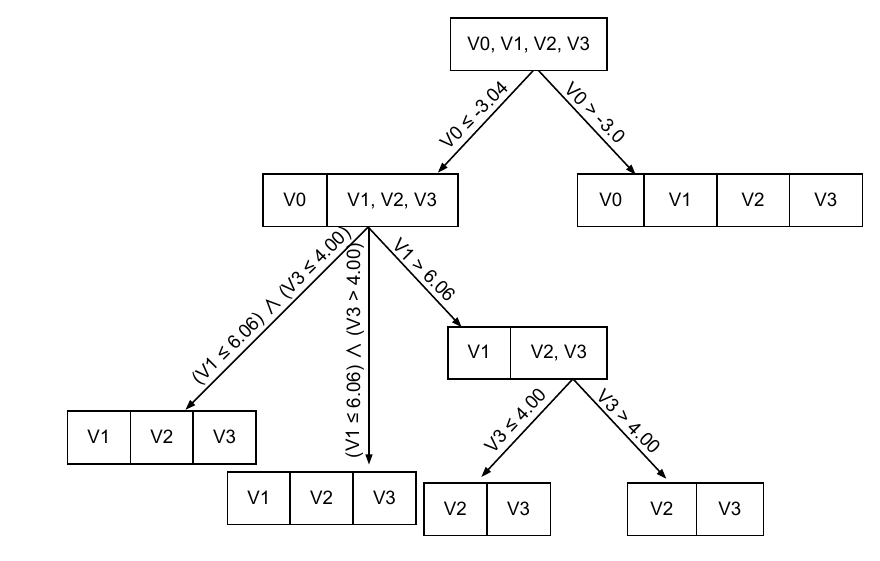}
    \caption{\small The sum-product network (left) trained on the synthetic data set and its corresponding CSI-tree extracted by \ex (right). Each edge in the CSI-tree corresponds to an edge from a sum node to a product node in the SPN. All other edges are collapsed. Clearly, the CSI-tree encodes all the CSIs represented in the sum-product network.}\label{fig:artificial_spn}
\end{figure*}

\begin{table*}[!t]
\centering
\caption{\small Summary statistics for the CSI rules extracted from SPNs by \ex and the association rules by Apriori algorithm. NP - $\#$ Product Nodes, NR - $\#$ Rules, MA - Mean Antecedent length, MC - the Mean Consequent length, and CR -  Compression Ratio.} 
\label{tab:summarystats}
\small
\begin{tabular}{lrrrrrrrrrrr}
\toprule
 & \multicolumn{1}{c}{SPN} & \multicolumn{3}{c}{All CSIs} & \multicolumn{4}{c}{Reduced CSIs} & \multicolumn{3}{c}{{Association Rules}} \\  \cmidrule(lr){2-2} \cmidrule(lr){3-5} \cmidrule(lr){6-9} \cmidrule{10-12}
Dataset  & NP & NR & MA & MC & NR & MA & MC & CR & NR & MA & MC \\
\midrule
Synthetic &  7 & 7 & 2.29 & 2.57 & \textbf{3} & 1.33 & 2.67 & 2.33 & 12 & 1.25 & 1.25 \\
Mushroom &  39 & 39 & 5.90 & 8.54 & \textbf{14} & 4.79 & 7.93 & 2.79 & 10704 & 2.87 & 2.43  \\
Plants &  342 & 342 & 9.60 & 9.61 & \textbf{23} & 6.22 & 7.09 & 14.87 & 1043 & 1.72 & 1.40 \\
NLTCS & 74 & 74 & 9.84 & 3.32 & \textbf{19} & 6.32 & 4.05 & 3.89 & 165 & 1.96 & 1.28 \\
MSNBC & 8 & \textbf{8} & 4.12 & 5.75 & \textbf{8} & 4.12 & 5.75 & 1.00 & 16 & 2.38 & 1.00\\
Abalone &  194 & 194 & 11.31 & 7.00 & \textbf{4} & 4.25 & 2.00 & 48.50 & 730 & 2.15 & 1.71 \\
Adult & 263 & 263 & 14.49 & 4.02 & \textbf{19} & 7.37 & 2.74 & 13.84 & 917 & 2.24 & 1.72 \\
Wine quality  & 236 & 236 & 12.45 & 6.76 & \textbf{5} & 3.60 & 2.60 & 47.20 & 337 & 1.99 & 1.56 \\
Car &  18 & 18 & 5.22 & 2.50 & \textbf{14} & 5.21 & 2.64 & 1.29 & 19 & 1.58 & 1.00\\
Yeast & 181 & 181 & 16.20 & 3.26 & \textbf{10} & 7.90 & 2.30 & 18.10 & 50 & 1.52 & 1.52 \\
nuMoM2b  & 104 & 104 & 10.60 & 2.33 & \textbf{31} & 6.55 & 2.19 & 3.35 & 21 & 1.29 & 1.14 \\
\bottomrule
\end{tabular}
\end{table*}
\begin{table*}
\centering
\caption{\small Summary statistics for the case where the instance function is inferred after training. The columns are the same as Table \ref{tab:summarystats}.}\label{tab:csi_i}
\small
\begin{tabular}{lrrrrrrrr}
\toprule
 & \multicolumn{1}{c}{SPN} & \multicolumn{3}{c}{All CSIs} & \multicolumn{4}{c}{Reduced CSIs}   \\
 \cmidrule(lr){2-2}  \cmidrule(lr){3-5} \cmidrule{6-9}
Dataset & NP & NR & MA & MC & NR & MA & MC & CR \\
\midrule
Synthetic & 7 & 7 & 2.29 & 2.57 & 3 & 1.33 & 2.67 & 2.33 \\
Mushroom & 39 & 39 & 5.90 & 8.54 & 13 & 5.08 & 8.38 & 3.00 \\
Plants &  342 & 342 & 8.33 & 9.61 & 32 & 4.62 & 6.72 & 10.69\\
NLTCS &  74 & 74 & 9.74 & 3.32 & 19 & 6.32 & 4.05 & 3.89 \\
MSNBC &  8 & 8 & 4.12 & 5.75 & 8 & 4.12 & 5.75 &  1.00 \\
Abalone & 194 & 157 & 9.61 & 6.85 & 8 & 5.75 & 2.00 & 19.63 \\
Adult &  263 &  244 & 11.27 & 3.89 & 12 & 6.17 & 3.08 & 20.33 \\
Wine &  236 & 235 & 9.18 & 6.78 & 6 & 3.50 & 2.50 & 39.17 \\
Car &  18 & 18 & 5.22 & 2.50 & 14 & 5.21 & 2.64 & 1.29 \\
Yeast & 181 & 181 & 13.06 & 3.26 & 10 & 6.60 & 2.30 & 18.10 \\
nuMoM2b & 104 & 98 & 9.64 & 2.29 & 35 & 6.97 & 2.17 & 2.80 \\
\bottomrule
\end{tabular}
\end{table*}
\begin{figure*}[!h]
    \centering
    \includegraphics[width=0.9\textwidth]{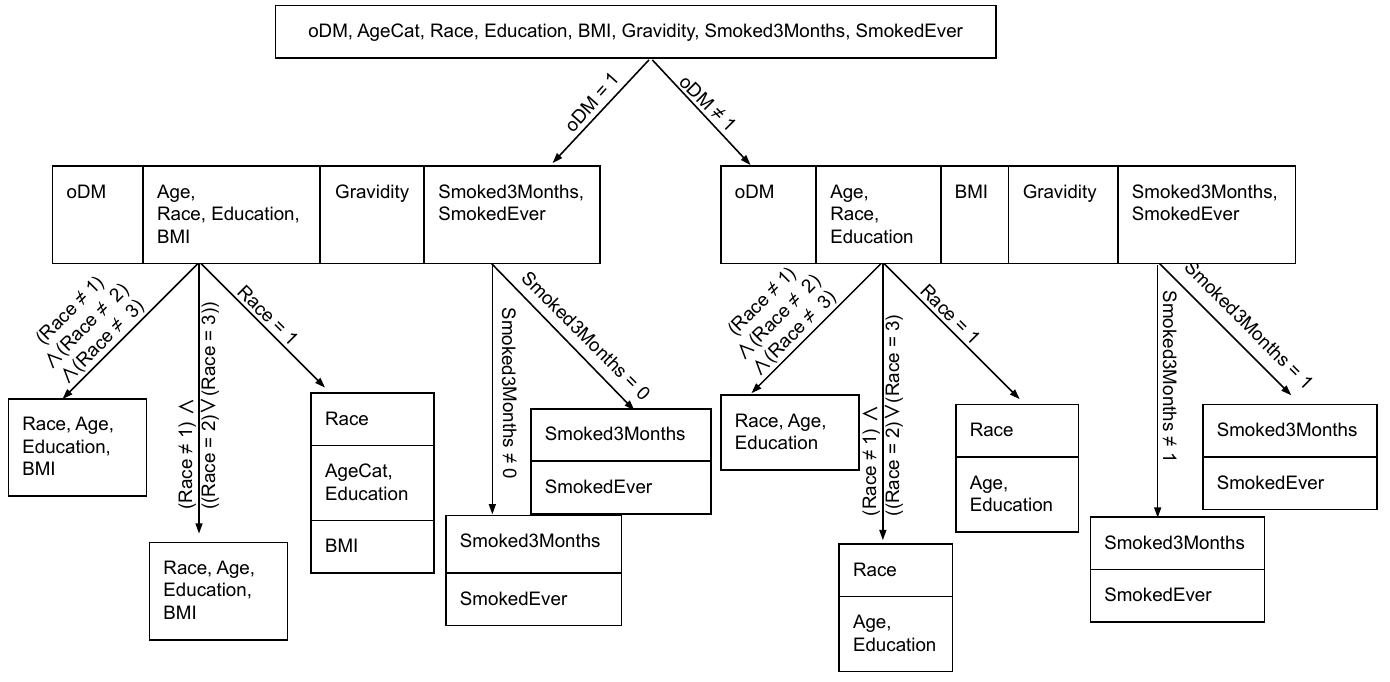}
    \caption{\small First two levels of the CSI-tree for the nuMoM2b dataset. Here, {\em oDM} is a boolean variable that represents whether or not the person has Gestational Diabetes. {\em Race} is a categorical variable having 8 categories namely, Non-Hispanic White (1), Non-Hispanic Black (2), Hispanic (3), American Indian (4), Asian (5), Native Hawaiian (6), Other (7), and Multiracial (8). {\em Smoked3Months} and {\em SmokedEver} are boolean variables representing tobacco consumption.}\label{fig:numom2b_csi}
    \vspace{-1em}
\end{figure*}
\textbf{Datasets:} We evaluated \ex on 11 datasets --  one synthetic, $9$ benchmark, and one {\bf real clinical study}. We generated the synthetic dataset by sampling 10,000 instances each from 3 multivariate Gaussians. We used $8$ datasets from the UCI repository and the National Long Term Care Survey (NLTCS, \cite{lowd2010learning}) data  from CMU StatLib (\texttt{http://lib.stat.cmu.edu/datasets/}). The 8 datasets from the UCI machine learning repository were Mushroom, Plants, MSNBC, Abalone, Adult, Wine quality, Car, and Yeast. In MSNBC and Plants, we used the rows which had at least two items present. For Wine quality, we concatenated the red wine and white wine tables. In addition, we used {\bf Nulliparous Pregnancy Outcomes Study: Monitoring Mothers-to-Be} (nuMoM2b, \cite{numom2b}) study data. Our subset has $8$ variables - {\em oDM, Age, Race, Education, BMI, Gravidity, Smoked3Months,and  SmokedEver}. Of these, oDM is the boolean representing  Gestational Diabetes ($0/1$).  We split each dataset into a train set having 75\% of the examples and a test set having the remaining 25\%. To ensure a balanced split, we stratified the splits on the target variable for the classification datasets. Table \ref{tab:data}, in the appendix. summarizes the datasets.

\textbf{Metrics:} We defined the following metrics on the CSIs - min\_precision ($mp$), min\_recall ($mr$), and n\_instances ($ni$). $mp$ and $mr$ of a CSI are the minimum values of precision and recall respectively for each of the decision rules that approximate the context. $ni$ of a CSI is the number of training instances in that context. We used thresholds on these metrics to obtain a reduced set of CSIs ($0.7$ for $mp$ and $mr$, and $5\times\texttt{min\_instances\_slice}$ for $ni$). We quantified this reduction as the {\em Compression Ratio (CR)}, which is the fraction of the total number of CSI rules in reduced set to the total number of rules. 
To compare association rules, we use mean antecedent length (mean $|A|$) and mean consequent length (mean $|C|$).
\vspace{-0.5em}
\subsection{Results}
\begin{wraptable}{R}{0.47\textwidth}
\caption{The Number of CSIs extracted by ExSPN from an SPN fit on each dataset (Total), The number of CSIs where at least 80\% of the datapoints in the context matched with the CSIs from the Bayesian Network(BN) (Correct) and the Ratio of correct CSIs (Ratio).  }
\label{tab:q1}
\centering
\begin{tabular}{llll}
\toprule
\textbf{Dataset} & Total & Correct & Ratio \\
\midrule
Earthquake & 12 & 10 & 0.83 \\
Cancer & 12 & 10 & 0.83 \\
Asia & 25 & 22 & 0.88\\
\bottomrule
\end{tabular}
\end{wraptable}
(Q1: Correctness) Table \ref{tab:summarystats} summarizes the SPNs, the full set of CSI rules extracted by \ex, and the reduced set of CSI rules. For each dataset in the table, the number of CSIs extracted by \ex(NR) {\bf is  exactly equal to} the number of product nodes of the SPN(NP). Figure \ref{fig:artificial_spn} shows the SPN learnt from the synthetic data and the CSI-tree extracted from that SPN using \ex. Clearly, the CSI-tree recovers all the CSIs encoded in the synthetic data. We further quantify how well the combination of SPN and \ex approximates the ground truth CSIs using data samples from BNs. Since the data is sampled from a BN, the ground truth CSIs can be obtained by converting the conditional distributions $P(X_i|X\setminus X_i) \forall X_i \in X$ to \emph{t}ree-{\em s}tructured {\em c}onditional \emph{p}robability \emph{d}istributions (Tree-CPDs). We use the ground truth CSIs to evaluate the CSIs extracted by \ex
Table \ref{tab:q1} summarizes the results of this evaluation. The ratios of the CSIs extracted by \ex to the ground truth is significantly high. Thus, Q1 is answered affirmatively.

(Q2: Compression) We can also infer from Table \ref{tab:summarystats} that filtering the CSI rules using min\_precision, min\_recall and n\_instances results in high compression ratios for {\bf all but the MSNBC dataset}. This is because the MSNBC dataset already had 8 rules and all of the rules satisfied the threshold conditions. Hence, Q2 is answered strongly affirmatively.

(Q3. Baseline) Table \ref{tab:summarystats} summarizes the association rules extracted from the data using the Apriori algorithm, and the mean confidence of the rules on the test set. Comparing the number of rules and the mean antecedent and consequent length from Table \ref{tab:summarystats} allows us to answer Q3. We can infer that while the CSI rules extracted by \ex are longer than the rules extracted by the Apriori algorithm, the set of CSI rules is much smaller.

(Q4. Are the explanations correct?) Figure \ref{fig:numom2b_csi} shows the first two levels of the CSI-tree extracted by \ex from the nuMoM2b dataset. The first split of the CSI-tree is on the target variable $oDM$. While the $BMI$ variable is independent of other variables when $oDM \neq 1$, it is dependent on $Age, Race, Education$ for the case when $oDM = 1$. 
These {\bf independencies are validated by our domain expert, Dr. David Haas}. This clearly demonstrates the potential of explaining a joint model such as SPN in a real clinically relevant domain. 
\section{Discussion and Conclusion}
We considered the challenging problem of explaining SPNs by defining a CSI-tree that captures the CSIs that exist in the data. We presented an iterative procedure for inducing the CSI-tree from a learned SPN by approximating the context induced by a product node using supervised learning. We then presented an algorithm for recovering an SPN that encodes the same CSIs as the original SPN from the CSI-tree thus establishing the correctness of the conversion. Our experiments in synthetic, benchmarks and a real clinical study demonstrate the effectiveness of the approach by identifying the correct CSIs from the data. As far as we are aware, this is the first work on explaining joint distributions using the lens of CSI.  Validating our method on more relevant clinical studies, allowing for domain experts to interact with our learned model, extending the algorithm to work on the broader class of distributions in general and TDPMs in particular, including more type of explanations, and finally, scaling the algorithm to large number of features remain interesting directions.
\vspace{-0.5em}
\section*{Acknowledgement}
The authors acknowledge the support by the NIH grant R01HD101246, AFOSR award FA9550-18-1- 0462 and ARO award W911NF2010224. KK acknowledges the support of the Hessian Ministry of Higher Education, Research, Science and the Arts (HMWK) in Germany, project “The Third Wave of AI”. DH acknowledges the support from the Eunice Kennedy Shriver National Institute of Child Health and Human Development (NICHD): U10 HD063037, Indiana University
\vspace{-0.5em}
\footnotesize
\bibliography{biblio,referencesSN}
\clearpage

\appendix
\section{Appendix}

\subsection{Details on \ex}
\begin{figure}[h]
    \sbox0{\includegraphics[width=0.45\textwidth]{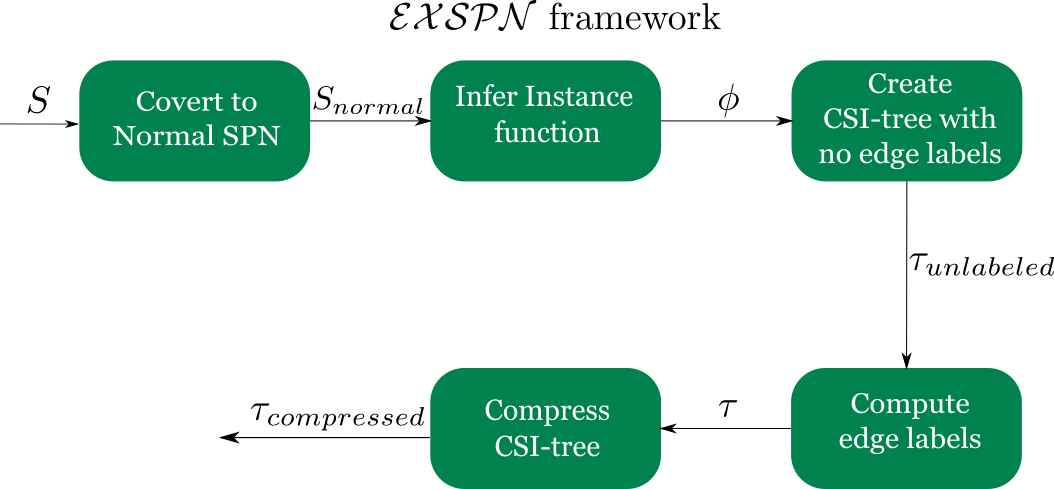}}
    \begin{minipage}{\wd0}
      \usebox0
      \caption{Flowchart illustrating the steps involved in converting an arbitrary SPN into a more interpretable CSI-tree.}\label{fig:flowchart}
    \end{minipage}
\end{figure}

The steps involved in converting an arbitrary SPN into a CSI-tree is illustrated in figure \ref{fig:flowchart}.

        \begin{algorithm2e}[H]
        \small
          \SetAlgoLined
          \caption{ComputeLabels}
          \label{alg:ComputeLabels}
          \SetKwInOut{Input}{input}\SetKwInOut{Output}{output}
          \Input{$\tau=(G, \psi, \chi, NULL), \da, \phi, \lambda$}
          \Output{Labeled CSI-tree $\tau=(G, \psi, \chi, \zeta)$}
          \For{Each edge $e(N_{from}, N_{to})$ in $G$}{
          Compute class labels $y$ \\
          $f=TrainModel(\da, \phi(N_{from}), \y)$ \\
          Compute a set of important features $I$ for $f$ \\
          $\zeta(e) = \wedge_{i \in I}i$ thresholded by $\lambda$
          }
          \textbf{return} $\tau=(G, \psi, \chi, \zeta)$
        \end{algorithm2e}
        \vspace{-1.5em}

\subsection{Proofs} \label{app:proofs}
\begin{algorithm2e}
\small
  \SetAlgoLined
  \caption{RetrieveSPN}
  \label{alg:RetrieveSPN}
  \SetKwInOut{Input}{input}\SetKwInOut{Output}{output}
  \Input{CSI-tree $\tau=(G_{csi}, \psi_{csi}, \chi, \zeta)$}
  \Output{$G_{normal}, \psi_{normal}$}
  \textbf{initialize:} $G_{normal}=G_{csi}, \psi_{normal}=\psi_{csi}$
  
  $st = [G_{normal}.root]$
  
  \While{$st$ is not empty}{
  $N_{current}=st.pop()$
  
  \uIf{$N_{current}$ has not been visited and is not the root node}{
  \uIf{$|\chi(N_{current})|=|\psi_{csi}(N_{current})|$}{
  Replace $N_{current}$ with $N_p$
  
  Append $|\chi(N_{current})|$ leaf nodes
  }
  \uIf{$|\chi(N_{current})|\neq|\psi_{csi}(N_{current})|$}{
  \For{$k$ in $\chi(N_{current})$}{
  \uIf{$|k|=1$}{
  Append $N_l$ s.t. $\psi(N_l)=k$
  }
  \uIf{$|k|\neq1$}{
  Append $N_s$ s.t. $\psi(N_s)=k$
  \For{$c$ in $ch(N_{current})$}{
  Append $N_p$ to $N_s$ if $\psi_{csi}(c)=k$
  }
  
  }
  }
  }
  }
  Add $ch(N_{current})$ to $st$
  }
  \textbf{return} $G_{normal}, \psi_{normal}$
\end{algorithm2e}

\begin{theorem}
 The CSI-tree $\tau=(G_{csi},\psi_{csi},\chi,\zeta)$, inferred from an SPN, $S_{normal}=(G_{normal}, \psi_{normal},$ $w_{normal},\theta_{normal})$, using \ex can infer $G'_{normal}$, and $\psi'_{normal}$ which encodes the same CSIs as $S_{normal}$.
\end{theorem}
\begin{proof}
  Algorithm \ref{alg:RetrieveSPN} that retrieves $G_{normal}$ and $\psi_{normal}$ associated with $S_{normal}$, given $\tau$ provides the proof. It performs DFS over $G_{normal}$ and identifies two main cases of $N_{current}$: \textbf{1.} A node where the length of the partition function $\chi(N_{current})$ is equal to length of the scope function $N_{current}$\textbf{[line 6]} \textbf{2.} A node where that's not the case\textbf{[line 10]}. In the first case, it replaces $N_{current}$ with a product node $N_p$ such that $\psi(N_p)=\psi_{csi}(N_{current})$ and adds $|\chi(N_{current})|$ number of leaf nodes. In the second case, it iterates over the subsets in $\chi(N_{current})$, adds a leaf node if that subset $k$ is a singleton set and adds an intermediate sum node $N_s$ for all other children associated with a subset $k$\textbf{[lines 10-17]}. Then, it returns computational graph $G$ and scope function $\psi$. Although this algorithm retrieves only $G_{normal}$ and $\psi_{normal}$ which define the structure of the SPN, the parameters of the SPN $w_{normal}$ and $\theta$ can also be retrieved by modifying \ex to produce a CSI-tree that stores these parameters in the leaf nodes of the CSI-tree alongside the scope $\psi_{csi}$ and the partition $\chi$. This modification is trivial and does not improve the interpretability. Hence, we do not consider it.
\end{proof}

\subsection{Details on Datasets and Hyperparameters}\label{app:data_hyper}
Tables \ref{tab:data} and \ref{tab:hyperparams} present statistics of the datasets and the hyperparameters used, respectively.

We generated the synthetic dataset by sampling 10,000 instances each from the following 3 multivariate Gaussians.
\renewcommand{\arraystretch}{0.8}
\begin{align*}
\small
&\mathcal{N}\begin{bmatrix}\begin{pmatrix}
2\\
2\\
2\\
2\\
\end{pmatrix}, 0.01\begin{pmatrix}
1 & 0 & 0 & 0\\
0 & 1 & 0 & 0\\
0 & 0 & 1 & 0\\
0 & 0 & 0 & 1\\
\end{pmatrix}\end{bmatrix},\\
&\mathcal{N}\begin{bmatrix}\begin{pmatrix}
-8\\
4\\
4\\
4\\
\end{pmatrix}, 0.01\begin{pmatrix}
1 & 0 & 0 & 0\\
0 & 1 & 1 & 1\\
0 & 1 & 1 & 1\\
0 & 1 & 1 & 1\\
\end{pmatrix}\end{bmatrix},\\
&\mathcal{N}\begin{bmatrix}\begin{pmatrix}
8\\
8\\
8\\
8\\
\end{pmatrix}, 0.01\begin{pmatrix}
1 & 0 & 0 & 0\\
0 & 1 & 0 & 0\\
0 & 0 & 1 & 1\\
0 & 0 & 1 & 1\\
\end{pmatrix}\end{bmatrix}
\end{align*}

\begin{table*}[!t]
\setlength\tabcolsep{3pt} 
\begin{minipage}{0.45\linewidth}
\small
\caption{Dataset details.}\label{tab:data}
\begin{tabular}{llrrrr}
\toprule
\textbf{Dataset} & Type & \multicolumn{1}{r}{$|\X|$} & \multicolumn{1}{r}{Train} & \multicolumn{1}{r}{Test}  \\
\midrule
Synthetic & Continuous & 4 & 22,500 & 7,500 \\
Mushroom & Categorical & 23 & 4,233 & 1,411  \\
Plants & Binary & 70 & 17,411 & 5,804  \\
NLTCS & Binary & 16 & 16,180 & 5,394  \\
MSNBC & Binary & 17 & 291,325 & 97,109  \\
Abalone & Mixed & 9 & 3,132 & 1,045  \\
Adult & Mixed & 15 & 33,916 & 11,306  \\
Wine quality & Continuous & 12 & 4,872 & 1,625  \\
Car & Categorical & 7 & 1,296 & 432 \\
Yeast & Mixed & 9 & 1,113 & 371  \\
nuMoM2b & Categorical & 8 & 8,832 & 388 \\
\bottomrule
\end{tabular}
\end{minipage}
\setlength\tabcolsep{3pt} 
\begin{minipage}{0.55\linewidth}
\small
\begin{flushright}
\caption{Hyperparameters(HP)}\label{tab:hyperparams}
  \begin{tabular}{lll}
\toprule
Component & HP & Value \\
\midrule
\multirow[t]{2}{*}{SPN} & rows & GMM \\
& $mis$ & \begin{tabular}[t]{@{}l@{}} 1\%\ of $|Train|$\\ 
$5\%$ of $|Train|$\\ 
1\%  of $|Train|$\\
\end{tabular} \\
\multirow[t]{3}{*}{DT} & max\_depth & 2 \\
 & $mid$ & 0.1  \\
 & class\_weight & balanced \\
\multirow[t]{3}{*}{\ex} & min\_precision & 0.7 \\
 & min\_recall & 0.7 \\
 & n\_instances & \begin{tabular}[t]{@{}l@{}}$5 \times mis$ \\ $mis$ (Synthetic) \end{tabular}\\
  \bottomrule
  \end{tabular}
\end{flushright}
\end{minipage}
\setlength\tabcolsep{3pt} 
\end{table*}

\subsection{Additional Experimental Results}
\renewcommand{\arraystretch}{1}
Table \ref{tab:loglik} presents the mean log-likelihood over the test set for the SPNs trained on each of the datasets.
      \begin{table}[!h]
        \caption{The mean log-likelihood over the test set (LL) for the SPNs trained on each of the datasets.}\label{tab:loglik}
        \begin{tabular}{lr}
        \toprule
        Dataset & LL \\ 
        \midrule
        Synthetic & 2.83 \\
        Mushroom & -8.98 \\
        Plants & -14.03 \\
        NLTCS & -6.3 \\
        MSNBC & -6.68 \\
        Abalone & 18.99 \\
        Adult & -5.52 \\
        Wine quality & -3.55 \\
        Car & -7.92 \\
        Yeast & 46.28 \\
        nuMoM2b & -6.92 \\ 
        \bottomrule
        \end{tabular}
        \end{table}

Table \ref{tab:aprioriconfig} presents the minimum support and minimum confidence parameters used for the apriori algorithm and the mean confidence of the association rules over the test set.

Tables \ref{tab:earthquake}, \ref{tab:cancer} and \ref{tab:asia} present all the CSIs extracted from the Earthquake, Cancer and Asia Bayesian Networks, and fraction of datapoints that match with the ground truth CSIs.
\begin{table}[!h]
\caption{The Minimum Support (MS) and Minimum Confidence (MC) parameters used for the apriori algorithm and the mean confidence of the association rules over the test set (TC).}\label{tab:aprioriconfig}
\begin{tabular}{lrrr}
\toprule
Dataset & TC & MS & MC \\ 
\midrule
Synthetic & 0.94 & 0.5 & 0.7 \\
Mushroom & 0.91 & 0.5 & 0.7 \\
Plants & 0.84 & 0.15 & 0.7 \\
NLTCS & 0.83 & 0.25 & 0.7 \\
MSNBC & 0.75 & 0.01 & 0.7 \\
Abalone & 0.87 & 0.25 & 0.7 \\
Adult & 0.85 & 0.5 & 0.7 \\
Wine quality & 0.86 & 0.5 & 0.7 \\
Car & 0.93 & 0.1 & 0.7 \\
Yeast & 0.9 & 0.5 & 0.7 \\
nuMoM2b & 0.87 & 0.5 & 0.7 \\ 
\bottomrule
\end{tabular}
\end{table}

\begin{table}[!h]
\caption{The CSIs extracted from data samples from the Earthquake Bayesian Network and the fraction of datapoints that matched with the ground truth CSIs}\label{tab:earthquake}
    \begin{tabular}{llr}
    \toprule
    Antecedent & Consequent & Matched \\
    \midrule
    (Alarm = 1) & (Earthquake, MaryCalls) & 1.0 \\
    (Alarm = 1) & (Earthquake, JohnCalls) & 1.0 \\
    (Alarm = 1) & (JohnCalls, MaryCalls) & 1.0 \\
    (Alarm = 1) & (Burglary, Earthquake) & 0.0 \\
    (Alarm = 1) & (Burglary, MaryCalls) & 1.0 \\
    (Alarm = 1) & (Burglary, JohnCalls) & 1.0 \\
    $\neg$(Alarm = 1) & (Earthquake, MaryCalls) & 1.0 \\
    $\neg$(Alarm = 1) & (Earthquake, JohnCalls) & 1.0 \\
    $\neg$(Alarm = 1) & (JohnCalls, MaryCalls) & 1.0 \\
    $\neg$(Alarm = 1) & (Burglary, Earthquake) & 0.0 \\
    $\neg$(Alarm = 1) & (Burglary, MaryCalls) & 1.0 \\
    $\neg$(Alarm = 1) & (Burglary, JohnCalls) & 1.0\\
    \bottomrule
    \end{tabular}
    \end{table}

    \begin{table}[!h]
    \caption{The CSIs extracted from data samples from the Cancer Bayesian Network and the fraction of datapoints that matched with the ground truth CSIs}\label{tab:cancer}
    \begin{tabular}{llr}
    \toprule
    Antecedent & Consequent & Matched \\
    \midrule
    (Cancer = 1) & (Pollution, Smoker) & 0.0 \\
    (Cancer = 1) & (Pollution, Xray) & 1.0 \\
    (Cancer = 1) & (Dyspnoea, Smoker) & 1.0 \\
    (Cancer = 1) & (Dyspnoea, Xray) & 1.0 \\
    (Cancer = 1) & (Smoker, Xray) & 1.0 \\
    (Cancer = 1) & (Dyspnoea, Pollution) & 1.0 \\
    $\neg$(Cancer = 1) & (Pollution, Smoker) & 0.0 \\
    $\neg$(Cancer = 1) & (Pollution, Xray) & 1.0 \\
    $\neg$(Cancer = 1) & (Dyspnoea, Smoker) & 1.0 \\
    $\neg$(Cancer = 1) & (Dyspnoea, Xray) & 1.0 \\
    $\neg$(Cancer = 1) & (Smoker, Xray) & 1.0 \\
    $\neg$(Cancer = 1) & (Dyspnoea, Pollution) & 1.0\\
    \bottomrule
    \end{tabular}
    \end{table}

    \begin{table}[!h]
    \caption{The CSIs extracted from data samples from the Asia Bayesian Network and the fraction of datapoints that matched with the ground truth CSIs}\label{tab:asia}
    \begin{tabular}{llr}
    \toprule
    Antecedent & Consequent & Matched \\
    \midrule 
    ((either = 1)$\land$$\neg$(asia = 1))$\lor$$\neg$(either = 1) & (bronc, smoke) & 0.0 \\
    ((either = 1)$\land$$\neg$(asia = 1))$\lor$$\neg$(either = 1) & (bronc, xray) & 1.0 \\
    ((either = 1)$\land$$\neg$(asia = 1))$\lor$$\neg$(either = 1) & (bronc, lung) & 1.0 \\
    ((either = 1)$\land$$\neg$(asia = 1))$\lor$$\neg$(either = 1) & (dysp, smoke) & 1.0 \\
    ((either = 1)$\land$$\neg$(asia = 1))$\lor$$\neg$(either = 1) & (dysp, xray) & 1.0 \\
    ((either = 1)$\land$$\neg$(asia = 1))$\lor$$\neg$(either = 1) & (dysp, lung) & 1.0 \\
    ((either = 1)$\land$$\neg$(asia = 1))$\lor$$\neg$(either = 1) & (dysp, tub) & 1.0 \\
    ((either = 1)$\land$$\neg$(asia = 1))$\lor$$\neg$(either = 1) & (bronc, tub) & 1.0 \\
    (((either = 1)$\land$$\neg$(asia = 1))$\lor$$\neg$(either = 1))$\land$$\neg$(lung = 1) & (smoke, xray) & 1.0 \\
    (((either = 1)$\land$$\neg$(asia = 1))$\lor$$\neg$(either = 1))$\land$$\neg$(lung = 1) & (tub, xray) & 1.0 \\
    (((either = 1)$\land$$\neg$(asia = 1))$\lor$$\neg$(either = 1))$\land$$\neg$(lung = 1) & (smoke, tub) & 1.0 \\
    (either = 1)$\land$(asia = 1) & (lung, tub) & 0.0 \\
    (either = 1)$\land$(asia = 1) & (bronc, lung) & 1.0 \\
    (either = 1)$\land$(asia = 1) & (dysp, lung) & 1.0 \\
    (either = 1)$\land$(asia = 1) & (dysp, xray) & 1.0 \\
    (either = 1)$\land$(asia = 1) & (tub, xray) & 1.0 \\
    (either = 1)$\land$(asia = 1) & (smoke, tub) & 1.0 \\
    (either = 1)$\land$(asia = 1) & (dysp, tub) & 1.0 \\
    (either = 1)$\land$(asia = 1) & (bronc, tub) & 1.0 \\
    (either = 1)$\land$(asia = 1) & (smoke, xray) & 0.56 \\
    (either = 1)$\land$(asia = 1) & (bronc, xray) & 1.0 \\
    (either = 1)$\land$(asia = 1) & (lung, smoke) & 1.0 \\
    (either = 1)$\land$(asia = 1) & (lung, xray) & 1.0 \\
    ((either = 1)$\land$(asia = 1))$\land$(bronc = 1) & (dysp, smoke) & 1.0 \\
    ((either = 1)$\land$(asia = 1))$\land$$\neg$(bronc = 1) & (dysp, smoke) & 1.0\\
    \bottomrule
    \end{tabular}
    \end{table}

\end{document}